\documentclass[letterpaper, 10 pt, conference]{ieeeconf}  

\IEEEoverridecommandlockouts                              

\usepackage{graphicx} 
\usepackage{amsmath} 
\usepackage{amssymb}  
\usepackage[linesnumbered,ruled,vlined]{algorithm2e}
\usepackage{booktabs}
\usepackage{cancel}
\usepackage{subcaption}
\usepackage{cite}
\usepackage{hyperref}
\usepackage{xcolor}

\newtheorem{lemma}{Lemma}
\newtheorem{remark}{Remark}
\newtheorem{assumption}{Assumption}

\newtheorem{definition}{Definition}

\title{\LARGE \bf
Combining Control Barrier Functions and Behavior Trees  \\ for Multi-Agent Underwater Coverage Missions
}

\author{\"Ozer \"Ozkahraman and Petter \"Ogren \\
KTH Royal Institute of Technology\\
\{ozero, petter\}@kth.se}

\begin{document}

\maketitle
\thispagestyle{empty}
\pagestyle{empty}

\begin{abstract}

Robot missions typically involve a number of desired objectives, such as avoiding collisions, staying connected to other robots, gathering information using sensors and returning to the charging station before the battery runs out.
Some of these objectives need to be taken into account at the same time, such as avoiding collisions and staying connected, while others are focused upon during different parts of the executions, such as returning to the charging station and connectivity maintenance.

In this paper, we show how Control Barrier Functions(CBFs) and Behavior Trees(BTs) can be combined in a principled manner to achieve both types of task compositions, with performance guarantees in terms of mission completion. We illustrate our method with a simulated underwater coverage mission.

\end{abstract}

\begin{figure}[t!]
\begin{subfigure}{\linewidth}
  \centering
  \includegraphics[width=.99\linewidth]{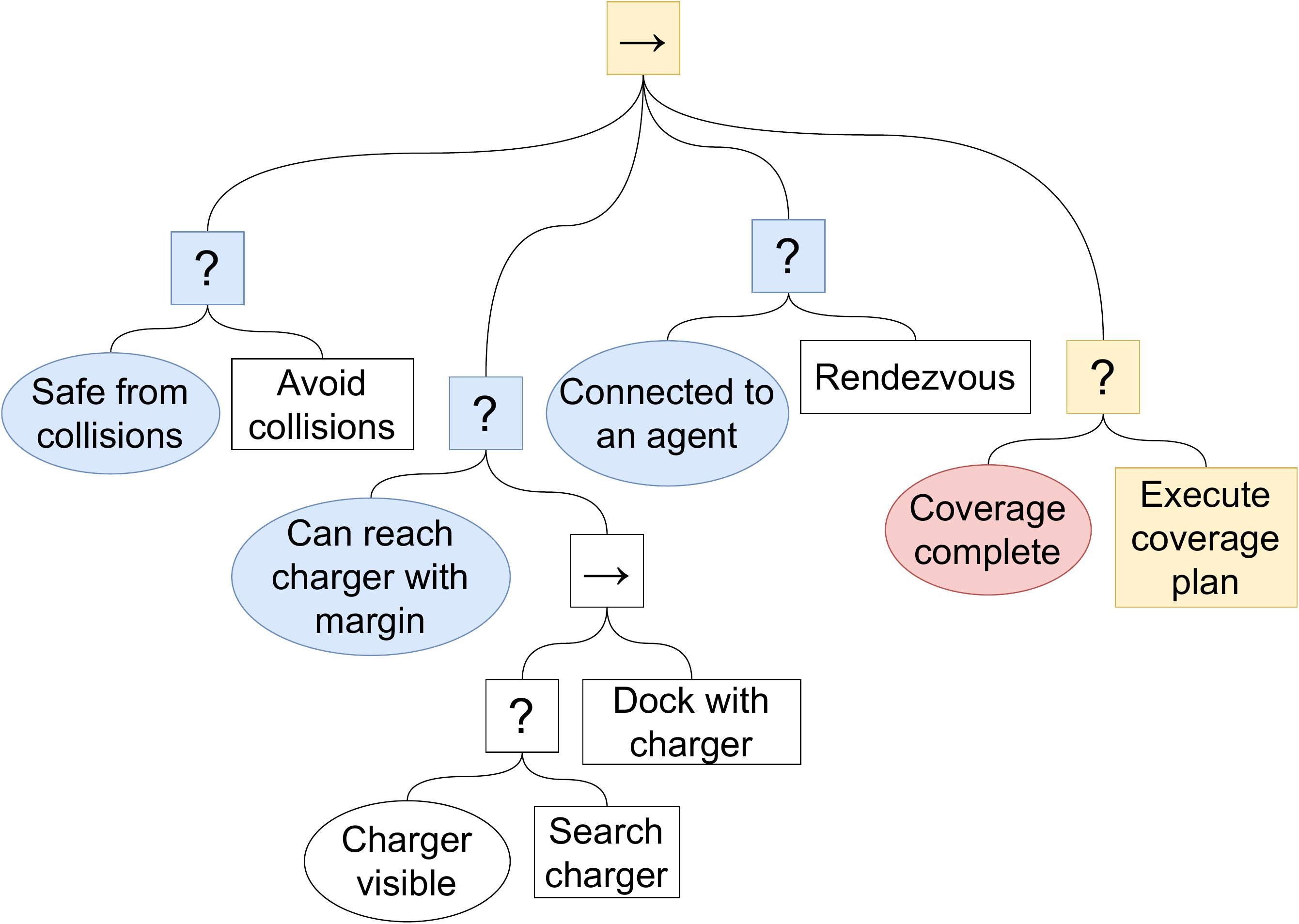}  
  \caption{Most conditions are satisfied and coverage is performed.}
  \label{fig:sub-first}
\end{subfigure}
\begin{subfigure}{\linewidth}
  \centering
  \includegraphics[width=.99\linewidth]{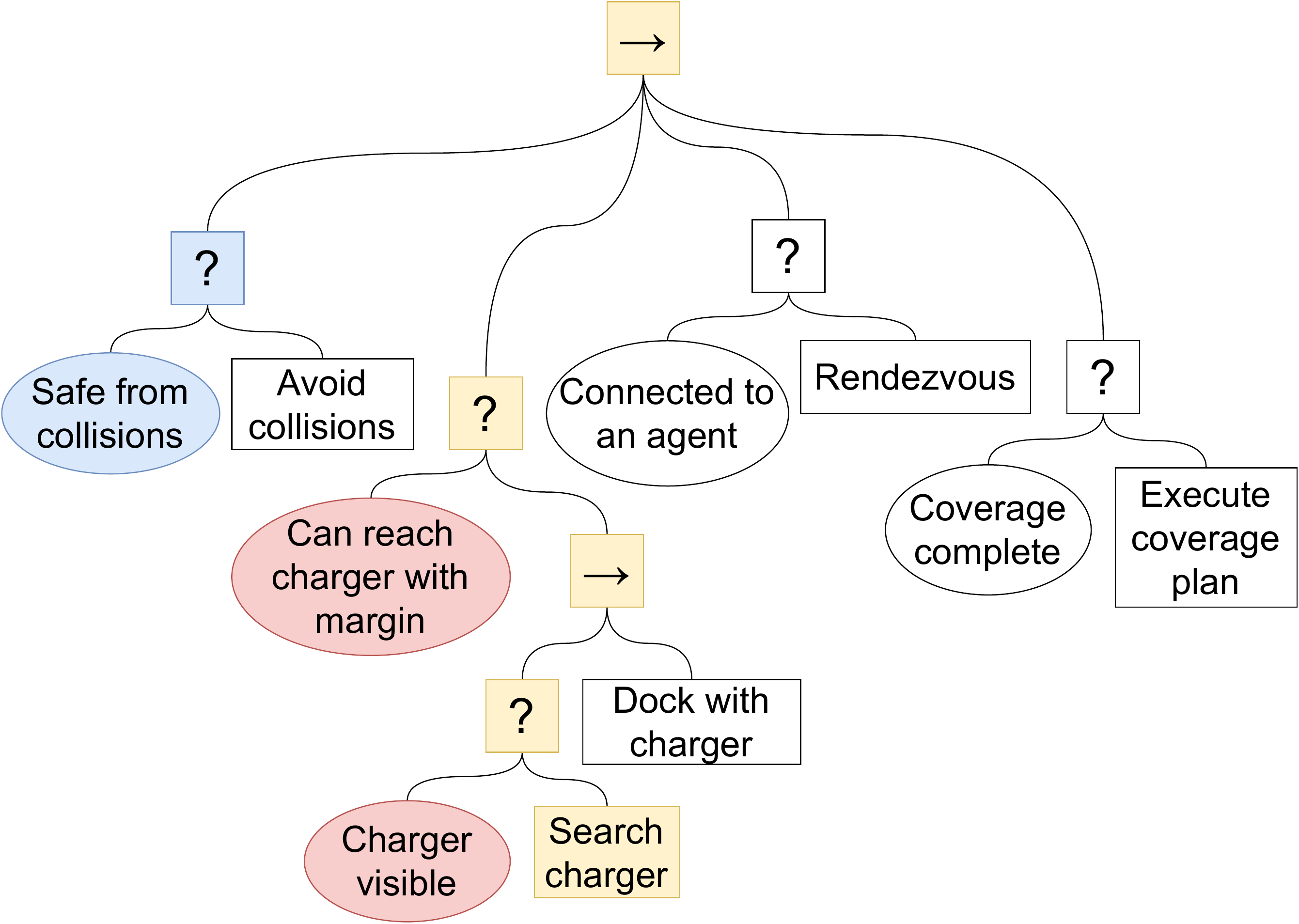}  
  \caption{The battery level is low and the robot is moving to the charger.}
  \label{fig:sub-second}
\end{subfigure}
\caption{The BT used for the multi-agent underwater coverage mission. The trajectories of the mission can be seen in Figure~\ref{fig:complex_sim}. In the figure, the satisfied conditions are blue, the unsatisfied conditions are red, and the  action that is currently executing is yellow.}
\label{fig:bt_ex1}
\end{figure}
\section{Introduction}
\label{sec:intro}

Behavior Trees (BTs) are a framework for designing and implementing sequential behavior switching in artificial intelligence (AI) applications  \cite{colledanchise_how_2017}.
They have been shown to generalize a set of earlier control architectures \cite{colledanchise_how_2017,colledanchise_how_2016}, such as the Teleo-reactive approach \cite{nilsson1993teleo}, Decision Trees, the Subsumption architecture \cite{brooks1990elephants} and Sequential behavior compositions \cite{burridge1999sequential}.
BTs were first used in games where some AI-controlled characters had to react to the player in increasingly complex ways \cite{isla2005gdc}.
In the domain of such games, the player is an unknown and dynamic part of the environment that might act in unforeseeable ways, so the AI-controlled character needs to react in real-time to what the player does.
This situation is also present in many robotics applications, where the environment has some part that cannot be predicted or controlled, such as the actions of nearby humans, or the location of unknown obstacles. 
Yet the robot is still expected to complete the given task.

Another similarity between robotics and games is that usually there are tasks that need to be completed in some order, such as first charging the battery and then moving an object. 
BTs provide a framework in which such sequential tasks can be encoded and executed in a structured and transparent manner.
 BTs can also provide reactivity, in the sense that when a previously completed task is undone by an external effect,  the planned sequence is interrupted to first repeat the previously completed task, and then continue with the sequence.

In addition to sequential tasks, robots often have parallel tasks that need to be fulfilled simultaneously, such as avoiding collisions and maintaining connectivity with other robots.
Control Barrier Functions (CBFs) \cite{ames2019control}, provide a framework in which such parallel tasks can be taken into account.
Using CBFs we can explicitly formulate constraints in the control space to guarantee invariance of some desired property.
CBFs have also been shown to be composable, such that multiple CBFs that handle different constraints can be combined using Boolean operators,  \cite{glotfelter2017nonsmooth}.

 The BT for our Autonomous Underwater Vehicle (AUV) application is shown in Fig~\ref{fig:bt_ex1}. 
We will use this BT throughout the paper as an example, and now provide an informal description of how it works, but leave the details for Section \ref{sec:related_work}.
The ultimate goal of this BT, which is run on several AUVs, is to jointly complete coverage of an area,
while at the same time also avoid collisions, stay within range of the battery charger and stay connected to at least one other agent. These four main objectives (conditions) are shown in the ovals in the top row of Fig~\ref{fig:bt_ex1}.

The conditions are ordered according to priority from left to right, with safe from collisions being the most important. Thus, only if safety is satisfied the battery level is considered, and only if those two are satisfied, rendezvous with a group member will be considered and so on. Thus, in Fig~\ref{fig:bt_ex1}(a) all the first three conditions are satisfied and coverage is executed, while in (b) the distance to the charger is too long, compared to the remaining battery, and the AUV ignores coverage and connectivity, but not safety, to leave the group and search for the charger. 

In the proposed approach, we will use CBFs for maintaining the conditions already met (blue in the figure), while striving to satisfy the next one (red). However, the approach is more complex than just maintaining a list of CBFs. First note that there are inherent conflicts in the mission, at some point, the battery level will force the AUV to stop doing coverage, so we need a mechanism for removing lower priority CBFs from the list of constraints. Furthermore the topology of the BT, in terms of subtrees, and the corresponding switching of tasks, has implications for what CBF to apply. When searching for the charger, only safety is considered while trying to get into visual range of the charger. But then, during docking, both safety and charger visibility are active CBFs. If the docking action accidentally brings the charger out of view, the AUV will switch back to searching for the charger before trying to dock once more. But this is the only action for which charger visibility is part of the active CBFs.

The  main contributions of this paper are as follows. 
We combine the BT structure, designed to handle sequential objectives, with CBFs, created to handle concurrent objectives, into a single framework. We furthermore show how to provide formal guarantees, in terms of goal satisfaction, in this framework.

The structure of this paper is as follows. First in Section~\ref{sec:related_work} we go through the relevant previous works, then in Section~\ref{sec:background} we outline the important parts of the two methods we make use of in the rest of the paper. In Section~\ref{sec:approach} we formally define the proposed combination of the two methods.
We apply the approach to an AUV problem in
 Section~\ref{sec:solution} and conclude our work in Section~\ref{sec:conclusions}.

\section{Related work}
\label{sec:related_work}

In this section we will discuss related earlier work on Behavior Trees, Control Barrier functions, and approaches to improving their performance.


\subsection{Behavior Trees}

One of the main gripes about BTs have been their convergence properties to a successful final state.
This has been addressed in 
 \cite{colledanchise_how_2017,paxton2019,rovida_extended_2017,rovida_motion_2018}. 
 In these papers, ideas from \cite{nilsson1993teleo} were extended to BT compositions. 
 These also considered the case of sequences, as a subset of all BT compositions, where each action meets the preconditions of the next action in the sequence.
 
 Other properties such as robustness (regions of attraction), safety (avoidance of some regions) and efficiency (convergence within upper time bounds) of BTs where analyzed in \cite{colledanchise_how_2017}. 
  The focus of \cite{colledanchise_how_2017} was on sequential objectives, but the case of concurrent objectives is mentioned, and the risk of switching back and forth between actions trying to achieve different objectives is noted. 
  The approach of combining CBFs and BTs proposed in this paper partially solves the problem noted in \cite{colledanchise_how_2017}.

In \cite{rovida_extended_2017,rovida_motion_2018} the core notation of BTs was extended
 by adding pre- and post-conditions to the Action nodes and thus removing Condition nodes. 
 This modified model is called \textit{extended Behavior Trees} (eBT) and it is meant to be interfaced with Hierarchical Task Network (HTN) planning. 

A concept closely related to BTs is the Robust Logical-Dynamical Systems proposed in \cite{paxton2019}. 
There, sequences of actions are ordered by their proximity to the goal, and concatenated in a way so that each action pushes the state towards the runnable conditions of the next.
A probabilistic analysis of convergence is presented, based on bounds on success probability of each action, as well as an ambitious implementation in a real robotic system.

\subsection{Control Barrier Functions}

The area of CBFs is currently receiving an increasing amount of attention \cite{borrmann2015control,wang2016safety,glotfelter2017nonsmooth,gurriet2018towards,notomista2018persistification,li2018formally,ames2019control}. 
In \cite{borrmann2015control}, a method to control a swarm of agents such that they all stay  within the safe set of states was proposed. 
The method takes into account relative velocities and accelerations, in order to ensure safety using CBFs. 
The authors present both physical experiments and simulations showing that the proposed method keeps robots safe from collision while achieving their tasks.

\begin{table*}[!h]
\hspace{1mm}
\caption{The four node types of a BT.}
\begin{center}
\begin{tabular}{|c|c|c|c|c|c|}
\hline
 \bf{Node type} & \bf{Symbol}& \bf{Succeeds} & \bf{Fails} & \bf{Running} \cr
\hline
 Fallback  &? & If one child succeeds & If all children fail &If one child returns running \cr
\hline
Sequence &$\rightarrow$ &If all children succeed & If one child fails &If one child returns running \cr
\hline
Action & (name)& Upon completion & When impossible to complete & During completion \cr
\hline
Condition & (name) & If true & If false & Never  \cr
 \hline
\end{tabular}
\end{center}
\label{tab:nodeTable}
\end{table*}%

A method for determining control in a multi-robot setting was proposed in \cite{wang2016safety}, for a situation where the robots might not have the motion models of their peers. 
The method takes into account bounded accelerations and satisfies the safety criteria while minimizing the effect on the desired control input.
The authors also extended the method to work in a distributed manner.

A way to combine several CBFs using Boolean logic was proposed in \cite{glotfelter2017nonsmooth}. 
The example used concerns a group of single integrator agents moving towards goal points while avoiding collisions with each other and stationary disc shaped obstacles.
The method shows that the $min$ and $max$ operators on barrier functions are analogous to $and$ and $or$ operators in Boolean logic.
The authors showed that these operations can be used to compose multiple CBFs to work in tandem, in order to satisfy a more complex goal. 
We base our method on this work and build upon it by adding the switching mechanisms of BTs.

A method based on CBFs that takes into account non-linear disturbances and uncertainty was presented in \cite{gurriet2018towards}.
The paper formulates the problem as an optimization problem to be solved in order to guarantee safe set invariance.
The method is illustrated using an inverted pendulum example.

A paper that is close in spirit to the AUV examples presented here is
\cite{notomista2018persistification}, where the charge level of the robot is included as part of its state, such that the forward invariance of this extended state ensures the persistence of the task.
The authors used CBFs to synthesize controllers to keep the robot from depleting its batteries.
A persistent coverage task is used as an example to show the success of the method in simulation and with mobile robots.
The work takes into account the dynamics of the robots energy usage and augments the state and dynamical model with it.
By rendering the augmented state forward invariant with the use of CBFs, the robot is kept charged.
The main difference between this work and our work is that our method is more general in terms of task switching, relying on the topology of the BT to identify the proper CBFs.

The authors of \cite{li2018formally} proposed a method to chain together multiple behaviours in order to complete a high level mission. 
They argued that single behaviours are good for low-level control, but they do not address missions with multiple steps.
They introduced a method called Finite-Time Convergent CBFs that guarantees that the terminal configuration of a behaviour overlaps with the initial configuration of the next behaviour.
They validate their claims on a team of mobile robots.
These ideas are similar to our method, but the work proposed here goes further by using the flexible BT structure to control the agent.

Composition of sequential and parallel constrained control of multiple agents have been studied in \cite{srinivasan2018control}.
Their method combines linear temporal logics with control barrier functions to achieve the desired behavior.
However, their method assumes that all obstacles are known, whereas our method does not have such an assumption and reacts on-line to newly discovered conditions.

Learning based approaches\cite{fan2019bayesian, cheng2019end, khojasteh2019probabilistic} have also been considered together with CBFs.
In these works, the CBFs act as a guardian to the learning system and keep it within known-to-be-safe areas. 
These works are concerned with the unknowns at the control level, where the full dynamics of the system might not be fully known, and strive to fill in this gap by learning.

To summarize, we note that convergence of BTs has been addressed, but only for sequential structures, and never using CBFs. 
Conversely, compositions of several CBFs have been studied, sometimes in a switching setting, but never using a BT to structure the switching. 
Thus, the combination of BTs and CBFs has not been investigated earlier in the literature.

\section{Background}
\label{sec:background}

In this section we will provide a brief background on the two core concepts being applied in this paper, CBFs and BTs.
\subsection{Control Barrier Functions}

Let a system be described by $\dot x = f(x,u)$, $x\in \mathbb{R}^n$, $u \in U \subset \mathbb{R}^m$.
The key idea behind Control Barrier Functions \cite{ames2019control} is to specify a function $h:\mathbb{R}^n \rightarrow \mathbb{R}$ such that the so-called \emph{safe set} $\mathcal{C}$ is characterized by: 
$$
\mathcal{C} = \{x: h(x) \geq 0\}.
$$
Then, given the system dynamics $f(x,u)$, if we choose controls $u$ inside the set
\begin{align}
\label{eq:CBF}
K = \{u\in U: \frac{dh}{dx}f(x,u) \geq - \alpha(h(x))\},
\end{align}
we are guaranteed to stay in the safe set $x \in \mathcal{C}$, see  \cite{ames2019control}. Above, $\alpha$ is a so-called class $ \mathcal{K}$ function, that is $\alpha: \mathbb{R}_+ \rightarrow  \mathbb{R}_+, \alpha(0)=0$ and $\alpha$ is strictly monotonic increasing, see Theorem 2 in \cite{ames2019control}.

\subsection{Behavior Trees}

 A BT is a directed tree, with the usual definition of nodes, edges, root, leaves, children and parents.
In a BT, each node belongs to one of the four categories listed in Table  \ref{tab:nodeTable}.
Leaf nodes are either \emph{Actions} or  \emph{Conditions}, while interior nodes are either \emph{Fallbacks} or \emph{Sequences}.

A detailed comparison between Finite State Machines (FSMs) and BTs can be found in \cite{colledanchise_how_2017}.
Here, we note that the actions of the BT correspond to the states of the FSM, while the switching decisions that are distributed in the states of the FSM are captured in the tree structure.
This structure is defined by the Fallbacks, Sequences and Conditions, of the BT. 
Furthermore the switching decision in the BT is revised every $\Delta t$ by sending a so-called \emph{tick} to the root node.

The tick starts at the root of the BT and progresses down to its children, in a particular order from left to right.
When a leaf is ticked, it returns either  \emph{Success, Running} or \emph{Failure} to its parent, which either ticks the next child, or forwards the return status to its parent.
We will now describe the different node types in more detail.

\emph{Fallback.\footnote{Fallbacks are sometimes also called Selectors.}} 
The Fallback node ticks its children in order until one of them returns \emph{Success} or \emph{Running}.
Any child returning \emph{Success} means that the Fallback also returns \emph{Success}.
Any child returning \emph{Running} means that the child needs more time to execute the task, in which case the Fallback also returns \emph{Running}.
If all children return \emph{Failure} the fallback also returns \emph{Failure}.
In other words, the Fallback node acts like a contingency plan, where the next child is ticked as a response to the failure of the previous child.

\emph{Sequence.} 
The Sequence node ticks its children in order until one of them returns \emph{Failure} or \emph{Running}.
If a child returns  \emph{Success}, the sequence immediately ticks the next child.
If a child returns \emph{Failure} or \emph{Running}, the Sequence node returns the same result as the child.
In other words, the Sequence node acts as a list of actions to run in succession, and quits when a \emph{Failure} occurs anywhere in that succession.

\begin{figure}[t]
\centering
\vspace{5pt}
\includegraphics[width=0.47\columnwidth]{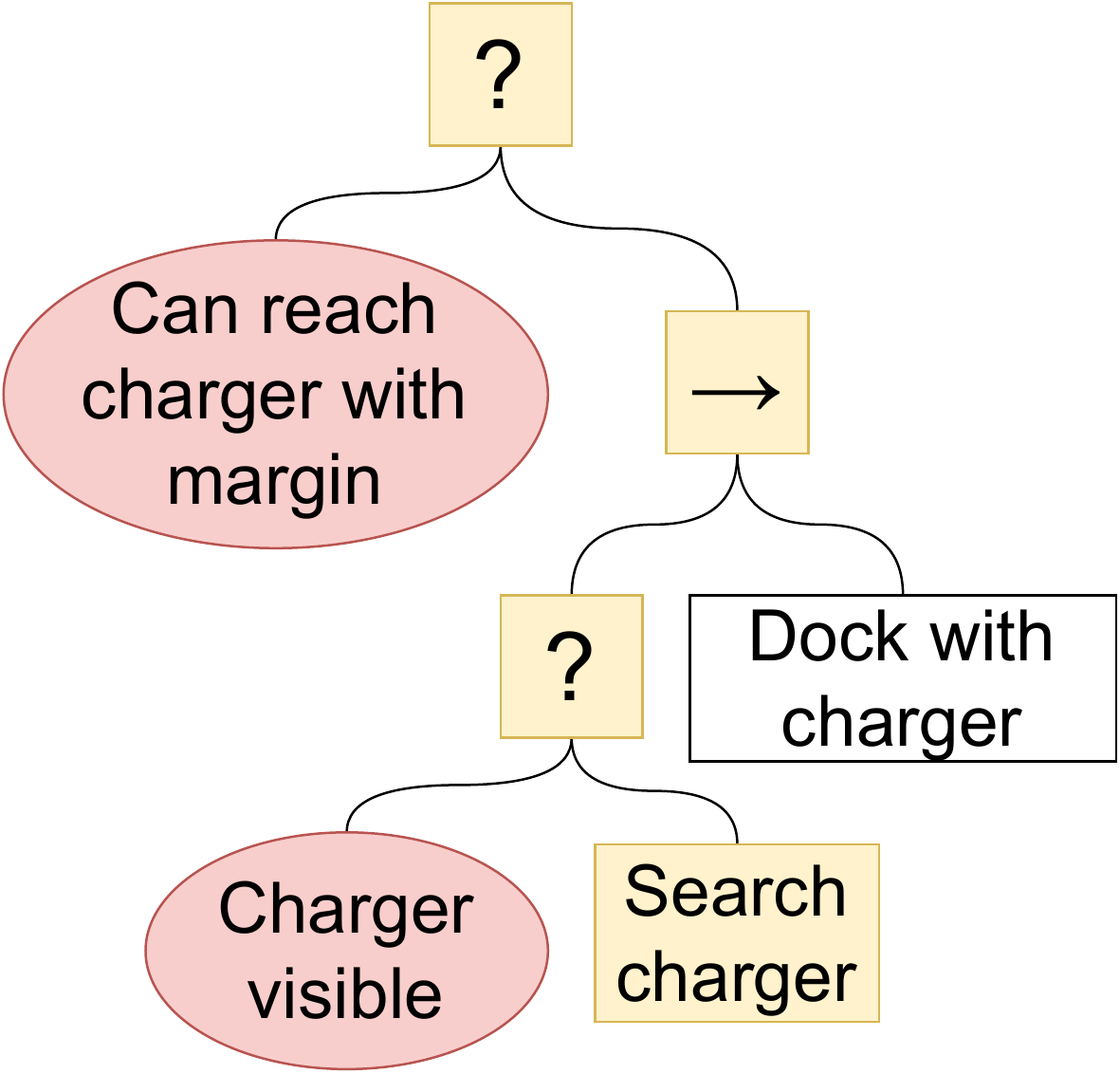}
\caption{A subset of the BT in Figure \ref{fig:bt_ex1}.}
\label{fig:complexbt_1_subset}
\end{figure}

To see how the components above can be combined into a larger structure we look at the BT in Figure \ref{fig:complexbt_1_subset}, which is a subset of  Figure \ref{fig:bt_ex1}. The action that is activated and get to control the AUV depends on the outcomes of the two conditions, which are summarized in Table~\ref{tab:BT_ex}.

\begin{table}[!h]
 \caption{Return status and action running as a function of the two conditions for the BT of Figure \ref{fig:complexbt_1_subset}.}
\begin{center}
\begin{tabular}{ | p{1.5cm} |p{1.5cm}| p{1.5cm}|p{2.4cm}|} 
 \hline
 \bf Can reach charger with margin &  \bf Charger visible  &  \bf BT returns &  \bf Action running \\ 
 \hline
 Success & Success & Success & (none)\\
 Success & Failure & Success & (none)\\
 Failure & Success & Running & Dock with charger\\
 Failure & Failure & Running & Search charger\\
 \hline
\end{tabular}
\end{center}
\label{tab:BT_ex}
\end{table}

The top node in Figure \ref{fig:complexbt_1_subset} is a Fallback, which means that as long as the first child returns success, it will return success itself, as can be seen in the two first rows of Table~\ref{tab:BT_ex}. In this case, the charger can be reached with a sufficient margin, and there is no need to worry about charging yet. Thus another part of the larger BT in Figure \ref{fig:bt_ex1} will be ticked. 

If a sufficient battery margin is not present, the top Fallback will tick its next child, which is a Sequence. 
This will in turn tick its first child which is a Fallback again. The second Fallback then ticks its first child, which is the second Condition: \emph{Charger visible}. If this Condition returns Failure the Fallback will tick \emph{Search charger}. If on the other hand, the Condition returns Success the Fallback returns Success and the Sequence above it ticks its next child, \emph{Dock with charger}. Thus we have covered all options described in Table~\ref{tab:BT_ex}. The whole BT in Figure \ref{fig:bt_ex1} can be analyzed in a similar way to see what actions are executed when. 

As illustrated above, the simple BT components can be combined to provide fairly rich switching strategies. Note also that the tree structure makes the design very modular,  it was quite straightforward to just pick a subtree and analyze it without knowledge of the rest of the BT.
For the interested reader,  a number of different design principles can be found in \cite{colledanchise2018behavior}.

Finally,
note that the actions represent continuous time controllers $\dot x = f(x,u_i)$, whereas the BT is executed in discrete time at intervals of $\Delta t$. Thus any switch between controllers $u_i$ can only occur at intervals of $\Delta t$. This results in a possible reaction delay in the switching with at most $\Delta t$, while at the same time guarantees that the number of switches in a finite time interval is bounded and solutions are well defined.

\section{Combining CBFs and BTs}
\label{sec:approach}
In this paper we are studying the problem of combining CBFs with BTs to create controllers that take a number of mission objectives into account, concurrently when needed, and sequentially when appropriate.
First we will present the nominal design, then a simple example illustrating the approach, then two extensions, to more complex BTs and more complex constraints.

\subsection{Nominal Design}
BTs can be constructed in many different ways, so first we define the general topology of BTs that we consider in this paper as follows.

\begin{definition}[Concurrent Goals BT (CG-BT)]
\label{def:CGBT}
A Concurrent Goals BT (CG-BT) is of the following form:

\emph{Sequence(Fallback($C_1$,$A_1$), 
$\ldots$, 
Fallback($C_N$,$A_N$))}, 

\noindent where $C_i$ is a condition, or a combination of conditions in the form of a BT, and $A_i$ is an action, or a BT including both actions and conditions.
\end{definition}{}

We use the name Concurrent Goals BT since it only returns Success if all the conditions (goals) $C_i$ are satisfied at the same time.

\begin{lemma}
\label{lem_convergence}
Given a Concurrent Goals BT, if the conditions $C_i$ are invariant, as in the returns do not change, under the actions to its right, $A_j, j>i$, and the execution of $A_i$ satisfies $C_i$ within some finite time $T$, then all $C_i$ will be satisfied within some time $TN$.
\end{lemma}{}

\begin{proof}
 We proceed by induction. First, if  $C_1$ is not satisfied, then $A_1$ will execute, and satisfy $C_1$ within time $T$. For the induction step, we note that if $C_k$ is not satisfied, while $C_i, i<k$ are satisfied, then $A_k$ will execute, keeping $C_i, i<k$ invariant, and satisfy $C_k$ within time $T$. Therefore we can conclude by induction that all $C_i$ will be satisfied within some time $TN$.
\end{proof}{}

\begin{remark}
Note that the assumptions of Lemma \ref{lem_convergence} are quite strong. In many examples, as will be shown below, they are not satisfied. But the result will still give important information on what designs to aim for and what performance can be expected from a given design.
\end{remark}{}

\begin{remark}[Conflicting goals]
\label{rem:conflicting}
Note that it is easy to construct conflicting goals. For example, the following design 
\emph{Sequence(Fallback($x >0$,\enspace$A_1$), Fallback($x < 0$,\enspace$A_2$))}, 
includes conditions (goals) that are impossible to achieve simultaneously. Clearly we cannot have $x > 0 \bigwedge x < 0 $, and it is impossible for $A_2$ to achieve $x < 0$ while keeping $x > 0$ invariant.
\end{remark}{}

For the AUV example, Table \ref{tab:cond} lists the set of conditions to satisfy for each action.
Note that this example is a bit more complex than Lemma  \ref{lem_convergence}, as it contains two nested BTs of the form CG-BT. In Section~\ref{subsec:recursive}
 below, we will describe how to produce this table from the topology of the BT.

\begin{table}[!h]
 \caption{Conditions to be kept invariant for the BT of Figure \ref{fig:bt_ex1}.}
\begin{center}
\begin{tabular}{ | l |p{5.5cm}| } 
 \hline
 \bf Action &  \bf Higher priority conditions to be kept invariant  \\ 
 \hline
 Avoid Collisions & (none) \\ 
 \hline
 Search charger & Safe from collisions \\ 
 \hline
 Dock with charger & Safe from collisions AND Charger Visible \\ 
 \hline
 Rendezvous & Safe from collisions AND Can reach charger \\ 
 \hline
 Execute Coverage & Safe from collisions AND Can reach charger AND connected\\ 
 \hline
\end{tabular}
\end{center}
\label{tab:cond}
\end{table}

In order to apply Lemma \ref{lem_convergence} above we need the conditions $C_i$ to be invariant under the actions to its right, $A_j, j>i$, and the execution of $A_i$ satisfy $C_i$ within some finite time $T$. To achieve such guarantees we use CBFs, and make the following assumption and definitions

\begin{assumption}
 Each condition $C_i:\mathbb{R}^n \rightarrow \{0,1\}$ can be formulated in terms of a CBF $h_i$, see (\ref{eq:CBF}), as follows 
  \begin{align}
  {C}_i &= (h_i(x) \geq 0)
  \end{align}
\end{assumption}

Given this assumption, we define the following sets of controls, 
where $K_i \subset U$ guarantees invariance of $C_i$, 
$\bar K_j \subset U$ guarantees invariance of all $C_i, i\leq j$ and
$\hat K_k \subset U$ guarantees invariance of a subset of $C_i, i\leq j$.
\begin{definition}
 Let 
 \begin{align}
K_i &= \{u \in U: \frac{dh_i}{dx}f(x,u) \geq - \alpha(h_i(x))\} \label{eq:CBFi} \\
\bar K_j  &= \bigcap_{i=1}^j K_i \\
\hat K_k &=\{ \bar K_j : j \leq k, \bar K_j \neq \emptyset \wedge (j = k \vee \bar K_{j+1} = \emptyset)  \}
\end{align}
\label{def:K}
\end{definition}

Note that $\bar K_j$ might be empty. This happens for example when the conditions are conflicting, see Remark \ref{rem:conflicting}, such as when \emph{Execute Coverage} requires the AUV to go north and \emph{Can reach charger} requires it to go south.
Thus, to address the problem of a possibly empty $\bar K_j $, we define $\hat K_k$ to be the intersection of the largest set of constraints $\mathcal{C}_j, j\leq k$ that is still  non-empty.

To apply Lemma \ref{lem_convergence} above we need controllers $A_i$ that satisfy $C_i$ within some finite time, while respecting the other constraints $C_j, j<i$. A set of different approaches to find such controllers can be envisioned. Ideally, a time optimal control approach to find a $u_i(t)$ that satisfies $h_i(x)\geq 0$ might be applied 

 \begin{align}
 \min_{u_i,x,T}& \int_0^T 1 dt   \label{eq:opt_ctrl} \\
\mbox{s.t.  } & \dot x = f(x,u_i) \nonumber \\
 & x(0)=x_0,\enspace x(T): h_i(x(T)) \geq 0 \nonumber \\
 & u \in \bar K_{i-1} \nonumber 
\end{align}

With lesser computational resources a reactive approach might also be used. Either one that locally maximizes the progression towards $C_i$, $h_i(x) \geq 0$
 \begin{align}
u_i = \mbox{ argmax}_u& \dot h_i(x,u)  \\
\mbox{s.t.  } & u \in \hat K_{i-1} \nonumber
\end{align}
or one that is close to some other control choice $k(x)$ that is designed to satisfy $C_i$, as in the CBF-QP of \cite{ames2019control},
 \begin{align}
u_i = \mbox{ argmin}_u& ||u-k(x)||^2  \label{eq:local_ctrl} \\
\mbox{s.t.  } & u \in \hat K_{i-1} \nonumber 
\end{align}

In this paper we will use the minimally disturbing version in (\ref{eq:local_ctrl}). 

Given the components above, we define the CBF-BT as follows.

\begin{definition}[CBF-BTs]
\label{def:CBFBT}
A Control Barrier Function Behavior Tree (CBF-BT) is a BT such that:
\begin{itemize}
    \item The BT is structured and indexed as a Concurrent Goals BT as defined in Definition~\ref{def:CGBT}. 
    \item Every Condition node $C_i$ is associated with a CBF $h_i$  and returns Success if $h_i(x)\geq 0$.
    \item Every action $A_j$ makes use of the $h_i$ according to (\ref{eq:local_ctrl}).
\end{itemize}
\end{definition}

With this definition we can formulate the second Lemma

\begin{lemma}
A CBF-BT will satisfy all constraints in finite time (See Lemma~\ref{lem_convergence}) if $\bar K_i=\hat K_i$ and 
\begin{align}
\mbox{max}_{u \in \hat K_{i-1}}& \dot h_i(x,u) \geq \epsilon > 0, \label{eq:local_ctrl_lemma} 
\end{align}
for some $\epsilon > 0$, throughout the execution.
\end{lemma}{}
\begin{proof}
First note that $\bar K_i=\hat K_i$ guarantees that there are no conflicts between the conditions, and the achieved constraints will be invariant \cite{ames2019control}. 
Then $\dot h_i(x,u) \geq \epsilon > 0$ implies that the next constraint will be satisfied in finite time.
Finally the result follows from Lemma \ref{lem_convergence}.
\end{proof}{}

\subsection{Simple example}

\begin{figure}[tb]
\centering
\vspace{5pt}
\includegraphics[width=0.99\columnwidth]{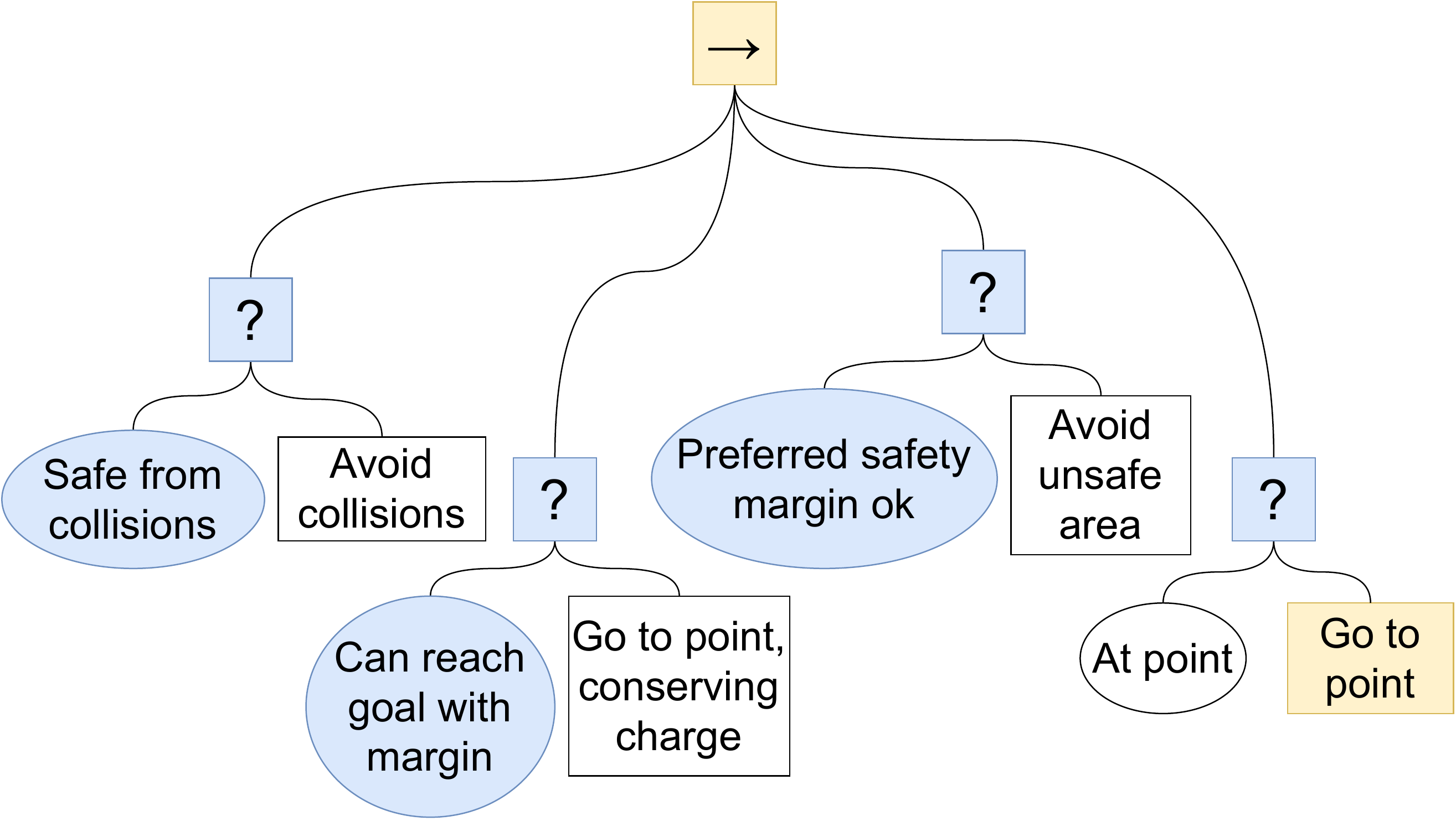}
\caption{A simple BT for getting to a goal location with desired margins on battery and obstacle clearance.}
\label{fig:simple_bt}
\end{figure}

To illustrate the approach so far we will look at the example in Figure \ref{fig:simple_bt} which is less complex than Figure \ref{fig:bt_ex1}. The BT has four concurrent objectives, in the following priority order: $C_1$: \emph{Safe from collisions}, $C_2$: \emph{Can reach goal with battery margin}, $C_3$: \emph{Preferred safety margin ok} and $C_4$: \emph{At point}, each paired with an action to achieve the condition. 
The conditions to be used as CBFs are given by Table \ref{tab:cond_simple}, 
and the detailed vehicle model can be found in Section \ref{sec:solution}.

\begin{table}[!h]
 \caption{Conditions to be kept invariant for the BT of Figure \ref{fig:simple_bt}.}
\begin{center}
\begin{tabular}{ | p{2cm} |p{5.5cm}| } 
 \hline
 \bf Action &  \bf Higher priority conditions to be kept invariant  \\ 
 \hline
 Avoid Collisions & (none) \\ 
 \hline
Go to Point, Conserving Charge &  Safe from collisions \\ 
 \hline
 Avoid Unsafe Area& Safe from collisions AND Can reach goal with battery margin \\ 
 \hline
 Go to point & Safe from collisions AND Can reach goal with battery margin  AND Preferred safety margin ok \\ 
  \hline
\end{tabular}
\end{center}
\label{tab:cond_simple}
\end{table}

\begin{figure}[htb]
\centering
\vspace{5pt}
\includegraphics[width=0.99\columnwidth]{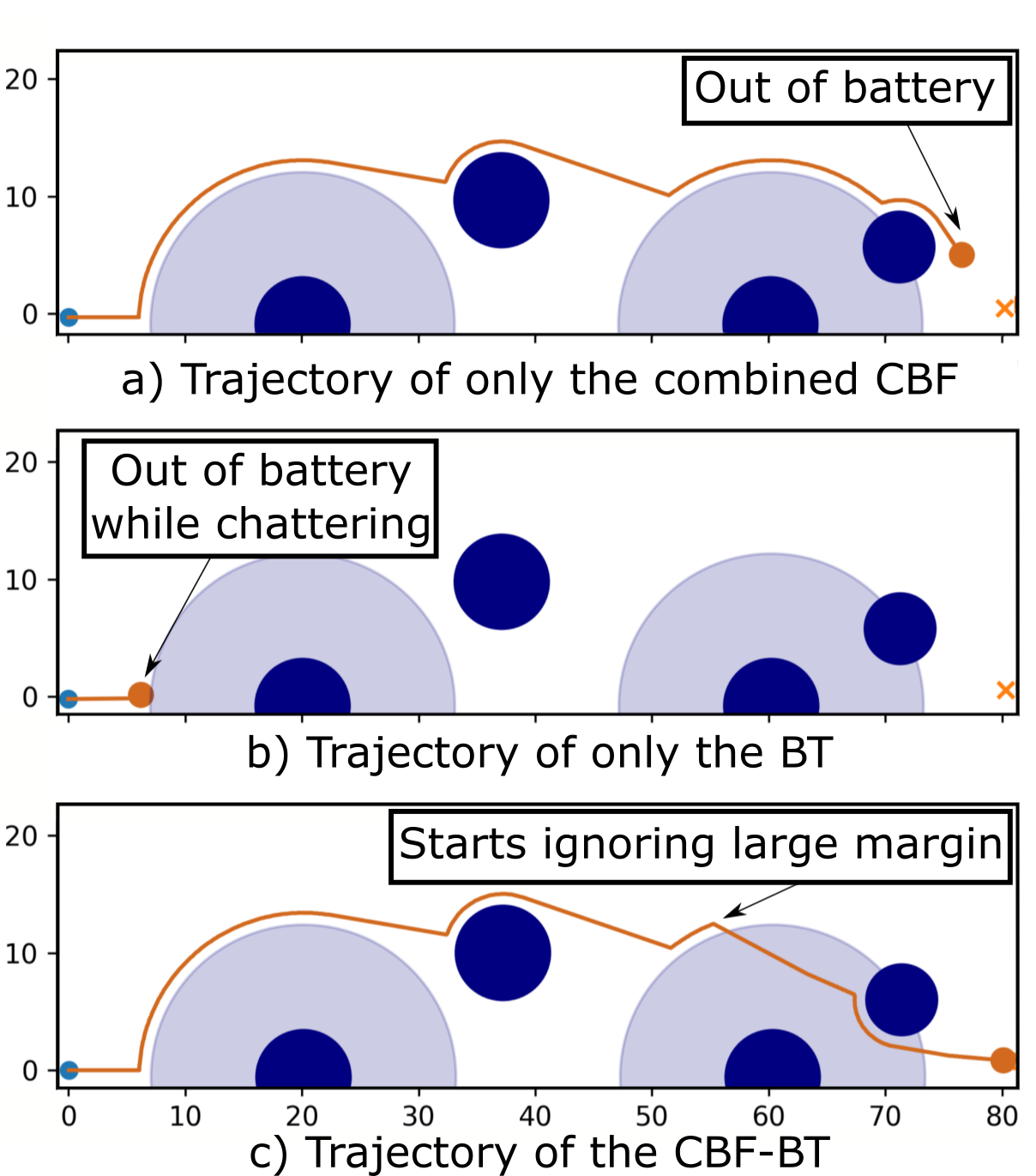}
\caption{The trajectories of the simple test case.}
\vspace{-8pt}
\label{fig:simple_sim}
\end{figure}

In Figure~\ref{fig:simple_sim}, three different simulated trajectories are shown.
The upper plot  shows that it is impossible to satisfy all constraints at the same time using a reactive controller. Trying to achieve $C_4$ while satisfying $C_1$ and $C_3$ results in the battery running out before reaching the goal. 

The middle plot shows the effect of running the BT in Figure  \ref{fig:simple_bt} without incorporating any constraints, i.e., ignoring $u_i \in \hat K_i$, in the actions. In this case, \emph{Goto point} would go straight towards the goal $C_4$, until hitting the light blue area and violating $C_3$. Then the BT would switch to  \emph{Avoid unsafe area} and move out of the area, then switch back to  \emph{Goto point}  and so on. A naive design leading to the battery running out while chattering.

The bottom plot shows the proposed design. Initially,  \emph{Goto point} is active, moving towards the goal while choosing controls in $\bar K_3 = \hat K_3$ satisfying all other constraints. Then, there is a conflict between \emph{Preferred safety margin ok} and \emph{Can reach goal with battery margin}, as the battery level is not high enough to reach the goal along the detour. This makes $\bar K_3 = \emptyset$ and thus $ \hat K_3= \bar K_2$, which effectively ignores \emph{Preferred safety margin ok} while still satisfying 
 \emph{Safe from collisions} and \emph{Can reach goal with battery margin}.

Thus, with several concurrent objectives, we have seen how the proposed approach provides graceful performance degradation to the CBFs when some constraints are conflicting, as well as reduces chattering of naive BT designs. However, we want to apply it to more complex designs exploiting the switching structure of the BTs more. 

\subsection{Recursive application of Lemma \ref{lem_convergence}}
\label{subsec:recursive}
We will now see how we can analyze more complex BTs by a recursive application of Lemma \ref{lem_convergence}, and get  the results of Table \ref{tab:cond}. Looking at Figure \ref{fig:bt_ex1} we see that it is a nested combination of two BTs of the CG-BT form, see Definition~\ref{def:CGBT}.

Applying Lemma  \ref{lem_convergence} at the top level we get  Table \ref{tab:cond_p1}. Here the recharging sub-BT is treated like any other action, and gets a set of constraints to satisfy, i.e. \emph{Safe from collisions}.  Applying Lemma  \ref{lem_convergence} at the sub-BT designed to take care of recharging we get Table  \ref{tab:cond_p2}, where we include the constraint given by the upper level. Note that Tables \ref{tab:cond_p1} and \ref{tab:cond_p2}, together give Table \ref{tab:cond}.

\subsection{More complex Constraints}
In this section we show how to transform a tree of conditions into a Boolean expression such as the one used for CBFs in \cite{glotfelter2017nonsmooth}.
Note that in Lemma 1, the conditions are either atomic conditions, or a tree of atomic conditions.

Since this mapping is only for condition nodes, the \emph{Running} return state does not need a representation, and we identify \emph{Success} with the Boolean \emph{TRUE} and 
\emph{Failure} with the Boolean \emph{FALSE}.
Let $c$ be any tree node and let $c_c$ represent its ordered list of children that does not include any actions.

\begin{definition}[$\FuncSty{Expand}$]
\begin{align}
    \FuncSty{Expand}^{}(c) =  \left\{
    \begin{array}{@{}l@{\thinspace}l}
        c, &\quad\text{$c$ is atomic} \\
        \bigvee_{c_i \in c_c} {\FuncSty{Expand}(c_i)}, &\quad\text{$c$ is Fallback} \\
        \bigwedge_{c_i \in c_c} {\FuncSty{Expand}(c_i)}, &\quad\text{$c$ is Sequence} \nonumber \\
    \end{array}
    \right.
\end{align}
\end{definition}
\vspace{2mm}
Using these definitions, and the approach proposed in \cite{glotfelter2017nonsmooth} we can convert a tree of constraints into a single CBF.

\begin{figure*}[t]
\vspace{3mm}
\centering
\includegraphics[width=\linewidth]{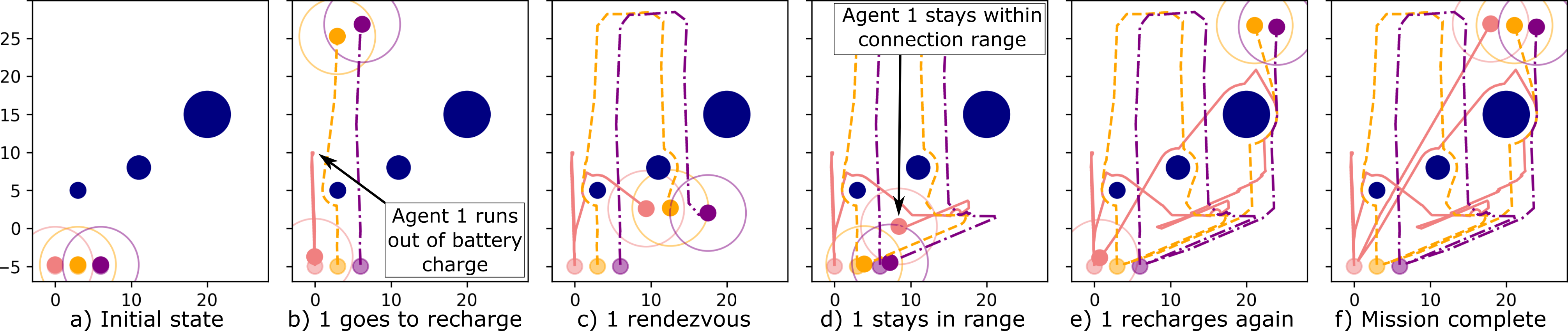}
\caption{The trajectories of the complex coverage mission (video available online$^2$).}
\label{fig:complex_sim}
\end{figure*}

\section{Persistent AUV coverage }
\label{sec:solution}

\begin{table}[!h]
\vspace{-5pt}
 \caption{Conditions to be kept invariant for the top level BT of Figure \ref{fig:bt_ex1}.}
\begin{center}
\begin{tabular}{ | l |p{5.5cm}| } 
 \hline
 \bf Action &  \bf Higher priority conditions to be kept invariant  \\ 
 \hline
 Avoid Collisions & (none) \\ 
 \hline
 Recharge sub-BT& Safe from collisions \\ 
 \hline
 Rendezvous & Safe from collisions AND Can reach charger \\ 
 \hline
 Execute Coverage & Safe from collisions AND Can reach charger AND connected\\ 
 \hline
\end{tabular}
\end{center}
\label{tab:cond_p1}
\end{table}

\begin{table}[!h]
 \caption{Conditions to be kept invariant for the Recharge sub-BT of Figure \ref{fig:bt_ex1}.}
\begin{center}
\begin{tabular}{ | l |p{2cm}|p{2.0cm}| } 
 \hline
 \bf Action &  \bf  Conditions from parent BT & \bf Conditions from this BT \\ 
 \hline
 Search charger & Safe from collisions & (none) \\ 
 \hline
 Dock with charger & Safe from collisions & Charger visible \\ 
 \hline
\end{tabular}
\end{center}
\label{tab:cond_p2}
\vspace{-3pt}
\end{table}

The problem considered is a set of AUVs that are performing an underwater coverage task in order to check the seabed for anomalies. 
The area to be covered is thought to have no obstacles, and a lawnmower pattern plan is made.
The AUVs are expected to always have a connected buddy for redundancy.
The AUVs run on battery, and can recharge in seabed mounted recharging stations. They can furthermore communicate with other AUVs within a given communication radius $r_c \in \mathbb{R}_+$ and sense the seabed with a given range $r_s\in \mathbb{R}_+$. The AUVs are modeled as kinematic points with bounded velocity and limited battery,
\begin{align}
    \dot x_i &= u_i, \\
    \dot b_i &= -k_b||u_i||,
\end{align}
where $x_i,u_i \in \mathbb{R}^2$,
are the position and control input respectively, $||u_i||\leq v_{\max}, ||u_i||\leq b_i$ .
$b\in [0, 100]$ is the battery charge level, and $k_b \in \mathbb{R}_+$ is the fuel efficiency of the AUV.

The AUVs should follow some safety constraints in order of importance:
\begin{enumerate}
    \item The AUVs must not get closer than some safety margin $m_s$ to each other and to static obstacles $o_k$.
    \item The battery level $b_i$ must be kept at a level such that the AUV can always reach the charging station position $q_i$ with some margin $m_b$.
    \item The AUVs are required to be connected to at least one other AUV while executing the plan for redundancy. 
\end{enumerate}

These requirements are formalized into conditions as:
\begin{align}
    ||x_i - x_j|| &\geq m_s, ~ \forall (i,j), ~ i\ne j \\
    ||x_i - o_k|| &\geq m_s, ~ \forall (i,k) \\
    ||x_i - x_j|| &\leq r_c, ~ \exists (i,j), ~ i\ne j  \\
    b_i &\geq ||x_i - q_i|| / k_b + m_b
\end{align}

In order to achieve the behavior we require, we design the CG-BT as shown in Figure~\ref{fig:bt_ex1}, with the corresponding CBFs of Table \ref{tab:cond}.
The actions in this tree satisfy Lemma~\ref{lem_convergence} by design. For example, the Action \emph{Rendezvous} will eventually cause the Condition \emph{Connected to an agent} to succeed by moving the agents towards each other, allowing a connection to be made.

It is clear that the charge margin condition will be eventually violated, independent of actions.
When this happens, the \emph{Execute coverage plan} action will be stopped, and the \emph{Charging subtree} will be run instead.
This switch constitutes a switch of tasks, from doing coverage to charging, and thus have different constraints applied as expected.
The charging subtree will then move the agent state to one where the charge margin is satisfied, allowing \emph{Rendezvous} and then \emph{Execute coverage plan} actions to be run in that order.


Having seen the desired effect on the simple test case, we simulate a short but complete coverage mission with three AUVs.
In Figure~\ref{fig:complex_sim}, the trajectories of all the AUVs are shown.
A video of the simulation is also available online\footnote{\url{https://youtu.be/TAi7zwfKw6o}}.
Similar to the simpler test case, the obstacles are unknown and no re-planning is done.
The CG-BT shown in Figure~\ref{fig:bt_ex1} is used on individual agents.
All of the constraints are in effect and checked throughout the simulation.
The coverage plan is set to be a simple lawnmower pattern with parallel turns and 3 lanes.
The safety margin $m_s$ is shown as filled disks for each agent.
The AUVs are identical and have the communication range $r_c$, shown as same-colored circles around each AUV in the figure.
The charging station for each AUV is at their individual initial positions.

In order to see the reaction of the system to unforeseen events, one of the AUVs (leftmost initial position, pink, solid curves) is started with only 30\% battery charge and the obstacles are scattered such that it affects one agent (middle initial position, yellow, dashed curves) more than the others.
This setup thus has multiple symmetry-breaking conditions in it such that all AUVs will be in different states throughout the simulation and must react accordingly to these unplanned circumstances.

As can be seen in Figure~\ref{fig:complex_sim}b, the first agent that started with less charge than the others had to return to its charging station soon after the mission started.
At this point, the connectivity constraint is not satisfied as expected.
Agent one later rendezvous with the group after its charged (c).

Shortly after the rendezvous with agent one, the other two AUVs run out of battery and return to their charging stations (d). 
Meanwhile agent one is running the \emph{Execute coverage plan} action which satisfies the connectivity constraint in a one-sided manner, by staying within communication range of the other agents. 
Even though agent one is trying to do coverage, it is essentially 'dragged' by the charging agents due to the CBFs.

Once the charging agents are done, we see that they all return to executing the coverage plan (e).
At this point, agent one, which has been staying as close to its coverage plan waypoint as it can without leaving the communication range, moves ahead of the other agents but runs out of battery again.
Agent one goes back to charge after the group resumes coverage and then rendezvous with the group a second time and completes the mission (f).
At this point, all agents have enough battery charge to return to their base stations and re-start the entire mission.

This simulation shows that the proposed method of CBF-BTs is capable of fulfilling the parallel tasks of avoiding collisions and keeping connectivity together with one of the sequential tasks of charging and following the coverage plan.
When the tasks become impossible to execute simultaneously, the CBF-BT is capable of prioritizing between the tasks as we observed when agents gave up on connectivity while going to re-charge.

\section{CONCLUSIONS}
\label{sec:conclusions}
In this paper, we proposed a general purpose reactive control method called Control Barrier Function Behavior Trees (CBF-BTs), that is able to take both concurrent and sequential objectives into account.
We have shown theoretical guarantees for goal satisfaction of CBF-BTs, and shown how they can be used to understand behavior even in cases where the underlying assumptions are not met.
The approach was illustrated with a collaborative AUV coverage mission.

\section*{ACKNOWLEDGMENT}
This work was supported by Stiftelsen for Strategisk Forskning (SSF) through the Swedish Maritime Robotics Center (SMaRC) (IRC15-0046).

\addtolength{\textheight}{-15cm}   

\bibliographystyle{unsrt}
\bibliography{ozer,cbf_papers,biblioShort}

\end{document}